\documentclass[10pt,twocolumn,letterpaper]{article}

\usepackage[pagenumbers]{cvpr} 
\usepackage{graphicx,amsmath,amsfonts,amssymb,caption,subcaption,multirow,overpic,textpos,makecell,url,booktabs,nicefrac,microtype,xspace,array}
\usepackage[table]{xcolor}
\usepackage[british, american]{babel}
\usepackage[pagebackref,breaklinks,colorlinks]{hyperref}

\usepackage[capitalize]{cleveref}
\crefname{section}{Sec.}{Secs.}
\Crefname{section}{Section}{Sections}
\Crefname{table}{Table}{Tables}
\crefname{table}{Tab.}{Tabs.}

\newlength\savewidth\newcommand\shline{\noalign{\global\savewidth\arrayrulewidth
  \global\arrayrulewidth 1pt}\hline\noalign{\global\arrayrulewidth\savewidth}}
\newcommand{\tablestyle}[2]{\setlength{\tabcolsep}{#1}\renewcommand{\arraystretch}{#2}\centering\footnotesize}
\renewcommand{\paragraph}[1]{\vspace{1.25mm}\noindent\textbf{#1}}

\newcolumntype{x}[1]{>{\centering\arraybackslash}p{#1pt}}
\newcolumntype{y}[1]{>{\raggedright\arraybackslash}p{#1pt}}
\newcolumntype{z}[1]{>{\raggedleft\arraybackslash}p{#1pt}}

\newcommand{\app}{\raise.17ex\hbox{$\scriptstyle\sim$}}

\newcommand{\x}{{\times}}
\definecolor{deemph}{gray}{0.6}
\newcommand{\gc}[1]{\textcolor{deemph}{#1}}

\definecolor{baselinecolor}{gray}{.9}
\newcommand{\baseline}[1]{\cellcolor{baselinecolor}{#1}}
\newcommand\Tstrut{\rule{0pt}{2.6ex}}

\def\cvprPaperID{*****} \def\confName{CVPR}
\def\confYear{2023}

\begin{document}

\title{Exploring Long-Sequence Masked Autoencoders}

\author{Ronghang Hu \qquad Shoubhik Debnath \qquad Saining Xie \qquad Xinlei Chen\\[4mm]
\texttt{\normalsize\url{https://github.com/facebookresearch/long_seq_mae}}\\[2mm]
}
\maketitle

\begin{abstract}
Masked Autoencoding (MAE) has emerged as an effective approach for pre-training representations across multiple domains.
In contrast to discrete tokens in natural languages, the input for image MAE is continuous and subject to additional specifications.
We systematically study each input specification during the pre-training stage, and find \emph{sequence length} is a key axis that further scales MAE.
Our study leads to a long-sequence version of MAE with minimal changes to the original recipe, by just decoupling the mask size from the patch size.
For object detection and semantic segmentation, our long-sequence MAE shows consistent gains across \emph{all} the experimental setups without extra computation cost during the transfer.
While long-sequence pre-training is discerned most beneficial for detection and segmentation, we also achieve strong results on ImageNet-1K classification by keeping a standard image size and only increasing the sequence length.
We hope our findings can provide new insights and avenues for scaling in computer vision.
\end{abstract}

\section{Introduction\label{sec:intro}}

Effectively processing data with rich structures is a long-standing and important topic in multiple fields of AI. It is \emph{aspirational}: \eg, it's exciting to build systems that can create proper high-resolution images from arbitrary language descriptions, or summarize any full-length novel into a succinct story plot.
It is also \emph{useful}: even for standard tasks such as image classification or object detection, providing more context or more details (\eg, by simply enlarging the inputs~\cite{tan2019efficientnet,Ghiasi2021}) to existing methods is widely accepted as a most reliable way to boost accuracies.

In fact, model developments can also be attributed to their ever-improving ability to capture signals within rich data. For computer vision, ConvNets~\cite{LeCun1989,Krizhevsky2012} supersede fully-connected networks with translation equivariance and pyramidal architectures -- both allowing better scaling \wrt input dimensions.
Recently, natural language processing (NLP) also witnessed a Transformer~\cite{Vaswani2017} revolution, where self-attention is used for `all-to-all' communications given a sequence of tokens. This design, in theory, can model all possible (short- or long-range) dependencies among tokens.
Inheriting this property, Vision Transformer (ViT)~\cite{Dosovitskiy2021} is quickly gaining popularity in computer vision, where a 2D input image is simply unfolded into fixed-sized patches for tokenization.

However, na\"{i}vely applying more powerful models to richer input data can incur problems. Besides efficiency~\cite{wang2020linformer}, a more serious concern is \emph{overfitting} -- large models tend to learn irrelevant intricacies and fail to generalize well when only a small number of training examples are available~\cite{Dosovitskiy2021}.
This issue can be further amplified by the `curse of dimensionality'~\cite{hastie2009elements} with expanded input sizes.

Fortunately, pre-training, especially variants of \emph{Masked Autoencoding (MAE)}~\cite{Devlin2019,he2021masked}, have risen as a domain-agnostic approach to reduce overfitting and scale models. For instance, BERT~\cite{Devlin2019} marked a paradigm shift to embracing gigantic models in NLP.
Extending its success to computer vision, MAE~\cite{he2021masked} again demonstrates strong model scalability by directly pre-training on raw pixels.
Unlike discrete text tokens, the richness of continuous visual signals (\eg images, video) depends not just on their content, but also on detailed specifications (\eg resolution, frame rate).
This presents a fresh opportunity for scaling that's barely explored in NLP.
Yet, given the standard practice~\cite{Dosovitskiy2021,Deng2009} followed by MAE, it is unclear how it will behave with different-sized images, particularly ones that depict complicated scenes and inherently need high-dimensional inputs.

In this paper, we study the input specifications of MAE. Two high-level choices are made for the rigorousness of our explorations: 1) \emph{decoupled} settings for pre-training and downstream transfers, which is possible with inputs but non-trivial for other specifications;\footnote{For example, model sizes are coupled for pre-training and downstream transfers, unless extra efforts are taken (\eg, distillation~\cite{hinton2015distilling}).} 2) as the relationship among MAE's input dimensions is deterministic, we study by fixing one dimension while jointly varying others.
These lead to our finding that a reasonably \emph{long} sequence length can yield meaningful gains \emph{without} incurring any extra downstream computation cost.
In other words, MAE pre-training \emph{by itself} also benefits from scaling sequence length.

A necessary design that enables our long-sequence MAE is to \emph{decouple} mask size from patch size. Masking patches individually works well for the original MAE~\cite{he2021masked}; but for a significantly longer sequence, it can bring about delicacies and degenerate the task \emph{even if} the same percentage of tokens are masked~\cite{xie2021simmim}. To maintain the difficulty level, our strategy is to simply `glue' nearby (\eg $2{\x}2$) patches for masking, so that multiple patches are jointly selected or deselected (see \cref{sec:approach:ours} and \cref{fig:random_vs_blocked_masking}). Without more advanced technical improvements as confounding factors, this design leads to a minimally-changed version of MAE and ensures the benefit of long-sequence pre-training is well-isolated.

For evaluation, we focus on object detection, instance segmentation, and semantic segmentation. Unlike iconic image classification~\cite{Deng2009}, these tasks operate on richer, real-world images that naturally translate to longer or larger feature maps~\cite{He2017,Xiao2018,li2022exploring}. Our long-sequence MAE \emph{consistently} helps performance, across \textbf{all} pre-training data (\eg COCO~\cite{Lin2014}, ImageNet-1K~\cite{Deng2009}, Open Images~\cite{kuznetsova2020open}, and Places~\cite{zhou2017places}), downstream benchmarks (\eg COCO, ADE20K~\cite{Zhou2019}, and LVIS~\cite{Gupta2019}), and models we have experimented \emph{without} any additional cost during the transfers. Moreover, \emph{better} scaling trend is empirically observed with larger models and scene-level images.
For the completeness of our study, we have also examined long-sequence MAE for ImageNet-1K image classification. Interestingly, while it brings noticeable gains with COCO pre-training, the gains are no longer discernible with ImageNet. This suggests our discovery indeed favors richer scenes as inputs. Nevertheless, if long-sequence inputs are supplied for both stages, we can achieve a similar top-1 accuracy to $448$-crop evaluation with standard $224$ crops. This again indicates sequence length -- not input resolution -- is a key factor for the final performance. 

We hope our methodology and findings can provide new insights and avenues for the broader scaling effort in computer vision. Code is made available.

\section{Related Work\label{sec:related}}

\paragraph{Masked autoencoders} are denoising autoencoders~\cite{Vincent2008} that aim to reconstruct complete signals given partial inputs. Their instantiation in NLP -- masked language modeling~\cite{Devlin2019,liu2019roberta,lan2019albert} -- has been proven tremendously successful.
Pioneered by earlier efforts~\cite{Vincent2010,Pathak2016,Chen2020c,Dosovitskiy2021}, many recent methods~\cite{Bao2021,he2021masked,wei2021masked,zhou2021ibot,xie2021simmim,dong2021peco,chen2022context,liu2022devil} have revisited this idea as a highly effective solution for visual representation learning. Notably, MAE~\cite{he2021masked} employs an explicit encoder-decoder architecture, and \emph{drops} (instead of \emph{`replaces'}~\cite{Devlin2019,Bao2021}) tokens for the heavier encoder. Such an efficient design makes it well-suited for our scalability study on sequence length.

\paragraph{\textnormal{We study} self-supervised learning} by \emph{decouple} pre-training and downstream transfers~\cite{li2022exploring}. Self-supervised learning holds the promise of \emph{scalability}, which stems from its unsupervised nature that saves labeling costs~\cite{Deng2009,zhou2017places}. On the other hand, multiple supervised benchmarks~\cite{Lin2014,Zhou2019} ensure \emph{diversity}, where task-specific designs~\cite{li2021benchmarking,li2022exploring} are often made to adapt pre-trained representations. Our attention is on pre-training; therefore we keep all the downstream hyper-parameters~\cite{li2021benchmarking,Xiao2018}, and especially input dimensions \emph{fixed} for our analysis. This not only helps a clean, \emph{scientific} understanding in contrast to prior studies that scale both~\cite{Chen2021a,Caron2021,li2021efficient}, but also offers a more efficient, \emph{practical} solution compared to scaling supervised transfers alone~\cite{xie2021simmim,he2021masked,dong2021peco}.\footnote{Speed is far more important for downstream transfers than pre-training -- a single pre-trained model will be fine-tuned many times, and be tested even more times.}

\paragraph{`High-resolution'} is a keyword closely related to `long-sequence'. Indeed, the ConvNet counterpart for long token sequences is high-resolution spatial feature maps. Typical ConvNet backbones~\cite{Krizhevsky2012,He2016} progressively shrink spatial dimensions of feature maps, which could be unfriendly to vision tasks that require accurate \emph{localization} (\eg detection, human pose estimation~\cite{Lin2014}). Many workarounds have been proposed. The simplest one is to increase image size, which has turned out to be a most reliable way to boost accuracies for numerous tasks~\cite{tan2019efficientnet,radosavovic2020designing,Lin2017,zhu2019deformable,Ghiasi2021,jiang2020defense} -- at the expense of more computation. More cost-effective solutions include: architecture designs~\cite{ronneberger2015u,Lin2017,wang2020deep}, strided convolutions~\cite{Chen2017,yu2015multi,zhu2019deformable}, cascading models~\cite{Ren2015,He2017,Cai2018}, \etc.

In the terminology of ViTs~\cite{Dosovitskiy2021,Touvron2021a}, high-resolution does not necessarily mean long-sequence -- the latter is also determined by the patch size used for tokenization. 
Then which one is the key? Scattered evidence suggests that sequence length is the key for supervised classification~\cite{beyer2022better} and detection transfer~\cite{chen2021simple} accuracies. This resonates with our findings and complements our systematic exploration for the pre-training stage \emph{without} changing downstream configurations.

\section{Approach\label{sec:approach}}

Our long-sequence MAE is a simple and minimally-changed extension of MAE~\cite{he2021masked}, which we summarize first.

\subsection{Background: MAE\label{sec:approach:background}}

\begin{figure}[t]
\centering
\vspace{-.1in}
\includegraphics[width=0.46\textwidth]{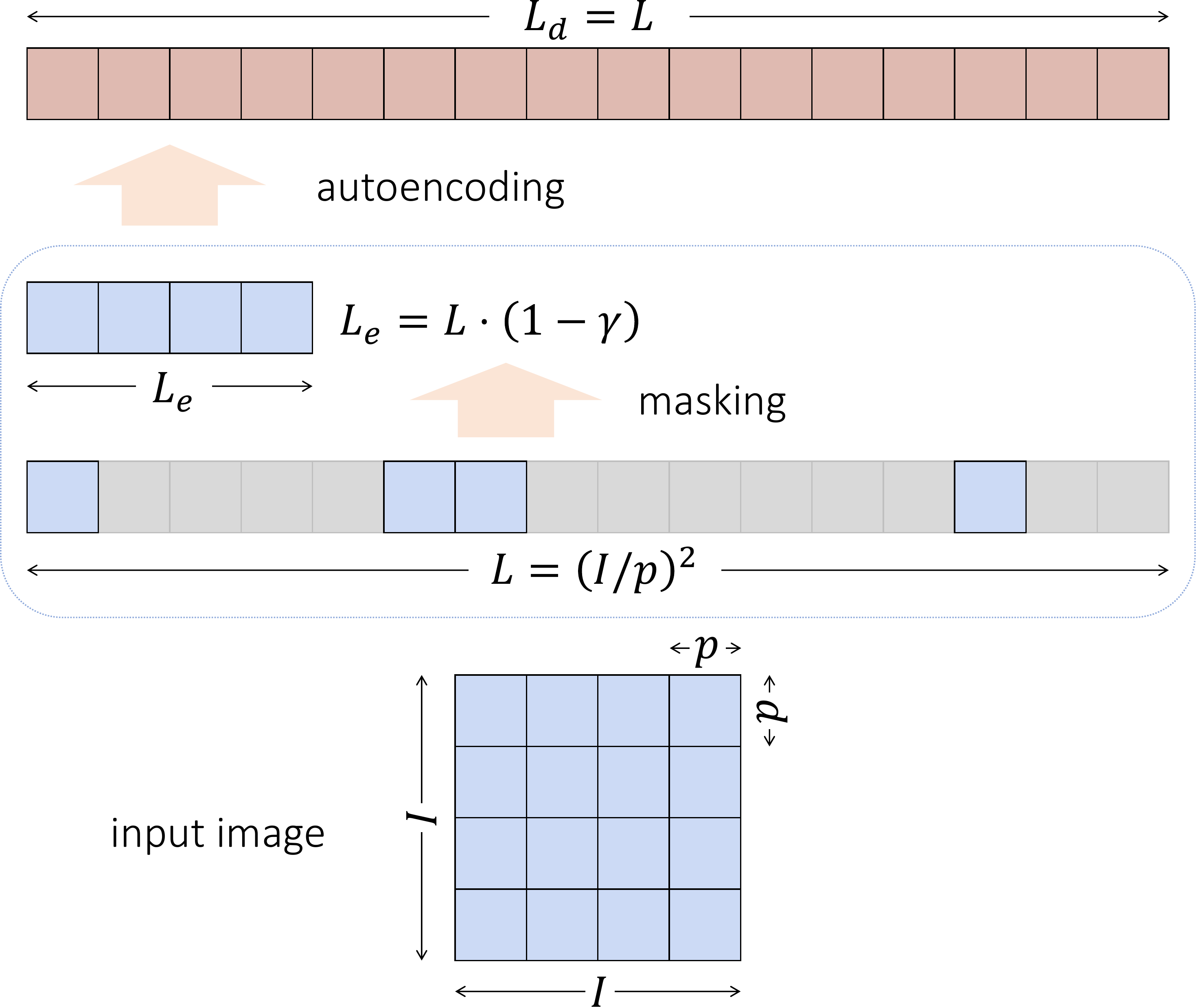}\\
\caption{\textbf{Input specifications for MAE}. An $I\x I$ input image is converted to a patch sequence of length $L=(I/p)^2$, according to the patch size $p\x p$. The encoder processes a random subset of $L_e=L\cdot(1-\gamma)$ patches based on the mask ratio $\gamma$, whereas the decoder produces all $L_d=L$ patches.}
\label{fig:approach}
\vspace{-.1in}
\end{figure}

Since our study focuses on the inputs (also extending to the outputs since it's an autoencoder), we introduce MAE by detailing the relevant specifications, as illustrated in \cref{fig:approach}.

\paragraph{Input specifications.} Following the practice of ViT~\cite{Dosovitskiy2021}, a 2D input image of size $I\x I$ is first unfolded into patches of size $p\x p$ for MAE. For simplicity, we use $I$ to denote image size and $p$ for patch size, as height and width are the same in all settings. 
Then the resulting 2D patch grid is flattened into a 1D patch sequence, leading to a total length of $L=(I / p)^2$ (see \cref{fig:approach} bottom). The default setup for MAE is $I=224$ and $p=16$, so $L=196$.

Next, a fixed number of patches are randomly masked according to a pre-defined ratio $\gamma$. Unlike BERT~\cite{Devlin2019}, masked patches are \emph{dropped} in MAE's encoder; and only visible patches remain. Therefore, the sequence length to the encoder, $L_e$, is computed as $L\cdot (1-\gamma)$. Coupled with a high mask ratio ($\gamma=0.75$), $L_e$ is only a quarter of $196$ (so $49$, see \cref{fig:approach} middle block). This makes MAE particularly efficient for high-capacity encoders, and ideal for our explorations on long-sequence pre-training (\eg compared to~\cite{Bao2021}).
The decoder still has a sequence length of $L_d = L$.

\paragraph{Autoencoding.} Given the input and output specifications above, a simple ViT-based architecture is adopted in MAE. The patch sequence to the encoder is first \emph{tokenized} via an embedding layer; then position embeddings are added and a \texttt{[CLS]} token is appended. The outputs of the encoder, after a projection layer that matches dimensions, are padded with $(L - L_e)$ \texttt{[MASK]} tokens (with position embeddings) and fed into the decoder. Finally, the output sequence of the decoder is used to predict the normalized pixel values~\cite{he2021masked} in the masked patches. $\ell_2$ loss is applied between the prediction of the decoder and the ground truth.

\paragraph{Downstream transfer.} The value of MAE pre-training lies in its capability to empower downstream transfer tasks. This includes supervised image classification, but more importantly object detection~\cite{li2021benchmarking} and semantic segmentation~\cite{Xiao2018} -- two prominent tasks in computer vision that require localized and holistic scene understanding. On these benchmarks, MAE shows solid improvements over other pre-training methods (\eg, supervised~\cite{Touvron2021a}, contrastive~\cite{Chen2021a}, or none), and scales well with model sizes (from ViT-B to ViT-L~\cite{Dosovitskiy2021}). We also focus on these tasks for our study.

\subsection{Methodology of study\label{sec:approach:method}}
We first discuss our high-level methodology to study input specifications for MAE. We highlight two aspects:

\paragraph{Decouple pre-training and transfer.} We want to isolate the effect of \emph{pre-training} changes, even though such effect can \emph{only} be measured by downstream tasks.
To this end, we decouple the input settings for pre-training and downstream transfers. That is, we \emph{only} change the input specifications for MAE, and \emph{fix} inputs for all the downstream tasks when studying the effect.
Note that it is non-trivial to make such a separation for model scaling (\eg conducted in MAE~\cite{he2021masked}, the same model size is used for both stages), whereas for input dimensions, we can easily change them due to the extensive weight-sharing used in modern model architectures. This differentiates us from prior work~\cite{Chen2021a,xie2021simmim}, both for cleaner scientific understanding and for faster practical deployment when scaling up, as discussed in \cref{sec:related}.

\paragraph{Fix one, vary two.} In common empirical studies, the \emph{degree of freedom} is usually equal to the number of variables (\eg network depth and width for model scaling~\cite{tan2019efficientnet,radosavovic2020designing}). In contrast, our main subjects of study -- image size $I$, patch size $p$ and sequence length $L$ have a deterministic relationship: $L=(I/p)^2$.
As a result, it is impossible to employ the typical strategy that \emph{vary one} variable while fixing the rest for studies.
Instead, we choose a \emph{reverse} strategy that fixes a specific dimension (\eg $I$) and lets the other two change \emph{jointly} (\eg $p$ and $L$). Going through all the combinations (shown in \cref{tab:ablation_study}), we arrive at a solid conclusion that sequence length is the \emph{key} for input scaling of MAEs.

\subsection{Long-sequence MAE\label{sec:approach:ours}}

Given our findings, there are multiple ways to instantiate long-sequence MAE. We simply adopt the following setting as default: $I{=}448$, $p{=}16$ and $L{=}784$. Compared to~\cite{he2021masked}, the sequence length is increased to $4\x$ (compute is also \app$4\x$), resulting from an enlarged image. As we will show experimentally, other settings (\eg $I{=}224$ and $p{=}8$) yield similar results as long as $L$ is maintained -- with one caveat:

\begin{figure*}[t]
\centering
\tablestyle{5pt}{1.2}
\begin{tabular}{cccccc}
\hspace{-1em}
\includegraphics[width=0.12\textwidth]{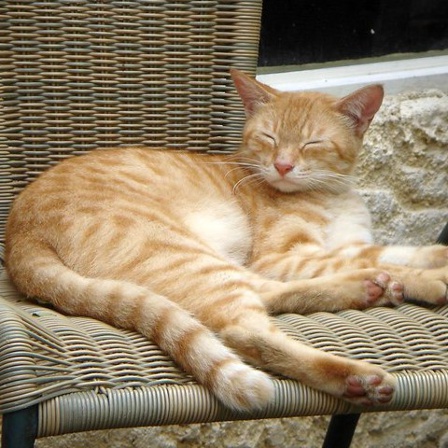} &
\includegraphics[width=0.12\textwidth]{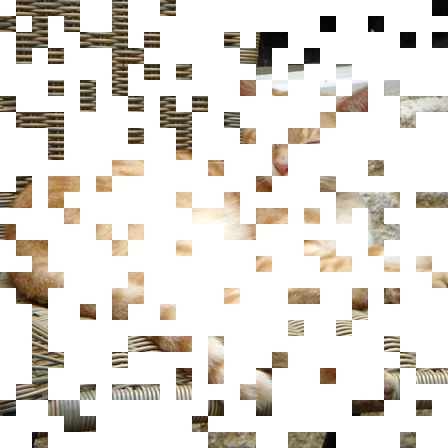} &
\includegraphics[width=0.12\textwidth]{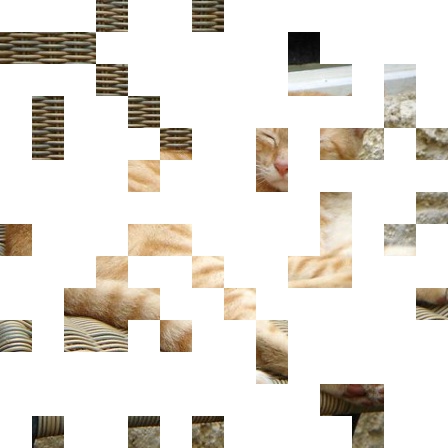} &
\includegraphics[width=0.12\textwidth]{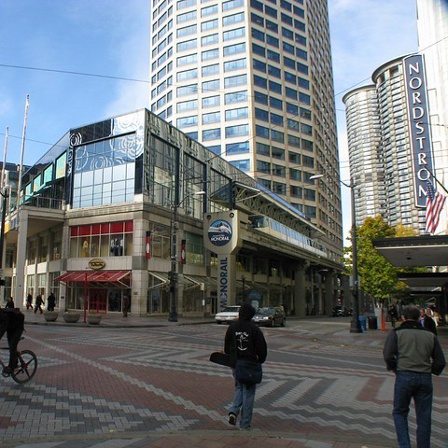} &
\includegraphics[width=0.12\textwidth]{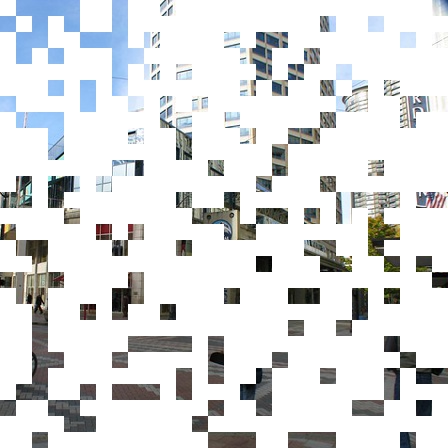} &
\includegraphics[width=0.12\textwidth]{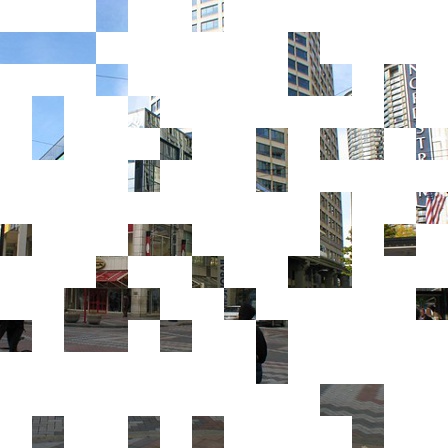} \\
\hspace{-1em}
image & $m=1\times 1$ & $m=2\times 2$ & image & $m=1\times 1$ & $m=2\times 2$
\end{tabular}
\caption{\textbf{Decoupled mask size} from patch size. Besides mask ratio ($\gamma=0.75$), mask size $m$ is another important factor that controls task difficulty, which was incidentally coupled with patch size in MAE~\cite{he2021masked}.
For long-sequence MAE, we reinstate $m$: if $m=1\x 1$, the model can mostly use near-by low-level texture details for reconstruction; $m=2\x 2$ renders a more semantically meaningful task.}
\label{fig:random_vs_blocked_masking}
\end{figure*}

\begin{table*}[t]
\tablestyle{10pt}{1.2}
\begin{tabular}{l|x{18}x{18}x{18}|x{18}x{18}x{18}|x{18}x{18}|x{28}}
& \multicolumn{3}{c|}{input specs} & \multicolumn{3}{c|}{masking strategy} & \multicolumn{2}{c|}{COCO} & ADE \\
\cline{2-7}
& $I$ & $p$ & $L$ & $m$ & $\gamma$ & $L_e$ & AP$^{b}$ & AP$^{m}$ & mIoU \\
\shline
BEiT~\cite{Bao2021} & 224 & 16 & 196 & $1\times 1$ & ~0.38\dag & 196 & 49.0 & 43.6 & 47.3 \\
\hline
MAE~\cite{he2021masked} & 224 & 16 & 196 & $1\times 1$ & ~0.75~ & 49 & 50.5 & 44.9 & 48.2 \\
ours, default & 448 & 16 & 784 & $2\times 2$ & ~0.75~ & 196 & \bf 51.7 & \bf 45.9 & \bf 50.8 \\
\end{tabular}
\vspace{-.1em}
\caption{\textbf{Baseline comparisons} on COCO and ADE20K with ViT-B. See \cref{sec:approach:background} and \cref{fig:approach} for definitions. For pre-training, BEiT~\cite{Bao2021} is \app$4\x$ in compute compared to MAE~\cite{he2021masked}; our default long-sequence MAE is also \app$4\x$, but deliver noticeable gains on both downstream tasks with the same input specifications ($I$, $p$ and $L$). 
(\dag: BEiT~\cite{Bao2021} uses another masking strategy and thus a different optimal mask ratio $\gamma$). 
\label{tab:baseline}}
\vspace{-.1in}
\end{table*}

\paragraph{Decoupled mask size.} Na\"{i}vely scaling up $L$ can degenerate MAE even if the mask ratio is maintained, due to the change in difficulty level (see explanations and examples in \cref{fig:random_vs_blocked_masking}).
The root cause here is that mask size and patch size are inherently \emph{two} variables -- the former controls the \emph{task} whereas the latter relates to the \emph{model} -- but they are incidentally set as a single one in the original MAE~\cite{he2021masked}.
Therefore, we reinstate mask size as a separate variable $m$, and \emph{jointly} select (or deselect) patches on the 2D patch grid for long-sequence MAE. For example, $m=2\x 2$ means the $2\x 2$ neighborhood is the basic \emph{unit} for masking. This is analogous to the commonly-used whole-word masking strategy in BERT~\cite{Devlin2019} and simplifies block-wise masking~\cite{Bao2021} where various-sized masks are encouraged.

\section{Experiments\label{sec:exp}}

We conduct our experimental analyses over a range of pre-training datasets and downstream tasks, starting with an ablation of the key factors in \cref{sec:exp:ablation_study} and a study on scaling trends of sequence length in \cref{sec:exp:seq_length}. We primarily focus on two pre-training datasets, namely COCO~\cite{Lin2014} and ImageNet-1K~\cite{Deng2009}, both of which have been used in previous self-supervised pre-training (\eg~\cite{Chen2020,el2021large}). We also explore pre-training on other common datasets including Open Images~\cite{kuznetsova2020open} and Places~\cite{zhou2017places} in \cref{sec:exp:other_pre-training_datasets}.

We adopt COCO object detection, instance segmentation, and ADE20K~\cite{Zhou2019} semantic segmentation, as our primary evaluation benchmarks for the pre-trained models. We also evaluate on other downstream tasks (LVIS~\cite{Gupta2019} instance segmentation) and detection architectures in \cref{sec:exp:simplefpn_coco_lvis}, where we observe consistent trends. Finally, we conduct experiments on ImageNet-1K classification (\cref{sec:exp:in1k_ft}).
We use Cloud TPUs for pre-training and GPUs for fine-tuning.

\subsection{Main results\label{sec:exp:ablation_study}}

We first conduct our main analysis on the input specifications of MAE, following the guidelines presented in \cref{sec:approach:method} and default setups in \cref{sec:approach:ours}. 

\paragraph{Setups.} By default, we pre-train ViT-B~\cite{Dosovitskiy2021} on COCO~\cite{Lin2014}, using the union of \texttt{train2017} and \texttt{unlabeled2017}\footnote{The \texttt{unlabeled2017} splits expands COCO \texttt{train2017} by ${\app}2\x$ and is confirmed beneficial, see \cref{sec:appendix}.} splits with a total of 241,690 images for 4,000 epochs (roughly 800 epochs on ImageNet-1K in training iterations as used in MAE~\cite{he2021masked}). Then we evaluate the pre-trained models by fine-tuning them on COCO object detection, instance segmentation, and ADE20K semantic segmentation as downstream benchmarks.

For COCO object detection and instance segmentation, we use a Mask R-CNN model~\cite{He2017} with a ViT-based detection backbone following~\cite{li2021benchmarking}, and fine-tune the pre-trained ViT model for 50 epochs (or roughly `4.1x') on COCO \texttt{train2017} annotations with the exact hyper-parameters and other details of~\cite{li2021benchmarking}, which are originally optimized for the MAE pre-trained models with $L=196$. For each pre-trained model, we report its box AP$^{b}$ and mask AP$^{m}$ on COCO \texttt{val2017}. Notably, the Mask R-CNN detection backbone always uses a ViT with patch size $16$ and image size $1024\times 1024$, and hence a fixed sequence length of $L=4096$ during detection fine-tuning for all pre-trained models.
When there is a mismatch in sizes, the ViT position embeddings are bicubic-interpolated to $L=4096$ following the practice in~\cite{li2021benchmarking}. The same is applied to patch embedding layers, where the weights are treated as 2D convolution filters and bicubic-interpolated when needed~\cite{li2022exploring}.

\begin{table*}[t]
\centering
\subfloat[
\textbf{Fix image size $I{=}448$}
\label{tab:ablate_image_size}
]{\begin{minipage}{0.27\linewidth}
\begin{center}
\small
\tablestyle{5pt}{1.2}
\begin{tabular}{cc|cc|c}
$p$ & $L$ & AP$^{b}$ & AP$^{m}$ & mIoU \\ 
\shline
64 & 49  & 44.0 & 39.8 & 35.0 \\
32 & 196 & 49.5 & 44.2 & 48.0 \\
\baseline{16} & \baseline{784} & \baseline{\textbf{51.7}} & \baseline{\textbf{45.9}} & \baseline{\textbf{50.8}} \\
\end{tabular}
\end{center}
\end{minipage}
}
\hspace{.5em}
\subfloat[
\textbf{Fix patch size $p{=}16$}
\label{tab:ablate_patch_size}
]{\begin{minipage}{0.39\linewidth}
\begin{center}
\small
\tablestyle{5pt}{1.2}
\begin{tabular}{cc|cc|c}
$I$ & $L$ & AP$^{b}$ & AP$^{m}$ & mIoU \\ 
\shline
112 & 49  & 47.3 & 42.1 & 42.2 \\
224 & 196 & 50.4 & 45.1 & 49.4 \\
\baseline{448} & \baseline{784} & \baseline{\textbf{51.7}} & \baseline{\textbf{45.9}} & \baseline{\textbf{50.8}} \\
\end{tabular}
\end{center}
\end{minipage}
}
\hspace{.5em}
\subfloat[
\textbf{Fix sequence length $L{=}784$}
\label{tab:ablate_sequence_length}
]{\begin{minipage}{0.27\linewidth}
\begin{center}
\small
\tablestyle{5pt}{1.2}
\begin{tabular}{cc|cc|c}
$I$ & $p$ & AP$^{b}$ & AP$^{m}$ & mIoU \\ 
\shline
224 & 8 & \textbf{51.7} & \textbf{46.0} & 50.5 \\
\baseline{448} & \baseline{16} & \baseline{\textbf{51.7}} & \baseline{45.9} & \baseline{\textbf{50.8}} \\
672 & 24 & \textbf{51.7} & 45.8 & 50.4 \\
\end{tabular}
\end{center}
\end{minipage}
}
\\[2mm]
\subfloat[
\textbf{Change mask size $m$}
\label{tab:ablate_mask_size}
]{\begin{minipage}{0.27\linewidth}
\begin{center}
\small
\tablestyle{5pt}{1.2}
\begin{tabular}{x{18}|cc|c}
$m$ & AP$^{b}$ & AP$^{m}$ & mIoU \\ 
\shline
$1\times 1$ & 50.5 & 45.0 & 47.8 \\
\baseline{$2\times 2$} & \baseline{\textbf{51.7}} & \baseline{\textbf{45.9}} & \baseline{\textbf{50.8}} \\
$4\times 4$ & 50.9 & 45.3 & 50.2 \\
\end{tabular}
\end{center}
\end{minipage}
}
\hspace{.5em}
\subfloat[
\textbf{Change lengths, $L_e$ and $L_d$}
\label{tab:ablate_enc_dec}
]{\begin{minipage}{0.39\linewidth}
\begin{center}
\small
\tablestyle{5pt}{1.2}
\begin{tabular}{ccc|cc|c}
$L_e$ & $L_d$ & $\gamma$ & AP$^{b}$ & AP$^{m}$ & mIoU \\ 
\shline
49 & 784~ & 0.9375 & 50.0 & 44.6 & 48.2 \\
\baseline{196} & \baseline{784~} & \baseline{0.75~} & \baseline{\textbf{51.7}} & \baseline{\textbf{45.9}} & \baseline{\textbf{50.8}} \\
196 & 196\dag & 0.75~ & 51.2 & 45.4 & 50.4 \\
\end{tabular}
\end{center}
\end{minipage}
}
\hspace{.5em}
\subfloat[
\textbf{ours \vs $4\x$ pre-trained MAE}
\label{tab:ablate_4x_schedule}
]{\begin{minipage}{0.27\linewidth}
\begin{center}
\small
\tablestyle{5pt}{1.2}
\begin{tabular}{l|cc|c}
& AP$^{b}$ & AP$^{m}$ & mIoU \\ 
\shline
\gc{MAE} & \gc{50.5} & \gc{44.9} & \gc{48.2} \\
MAE ($4\x$) & 50.0 & 44.5 & 48.5 \\
\baseline{ours} & \baseline{\bf 51.7} & \baseline{\bf 45.9} & \baseline{\bf 50.8} \\
\end{tabular}
\end{center}
\end{minipage}
}
\vspace{-.1em}
\caption{\textbf{Ablation studies} with COCO pre-trained ViT-B. We study: a-c) fixing one input dimension, and varying the other two among image size $I$, patch size $p$, and sequence length $L$; d) different mask sizes $m$; e) different encoder/decoder lengths $L_e$ and $L_d$, with optional changes to mask ratio $\gamma$; and f) \vs longer training schedule. COCO box AP$^{b}$, mask AP$^{m}$, and ADE20K mIoU are evaluated under the same transferring input sizes. Default settings are marked in \colorbox{baselinecolor}{gray}. (\dag: decoder length $L_d$ is first reduced to $196$ via a learned $2{\times}2$ convolution.) \label{tab:ablation_study}}
\vspace{-.1in}
\end{table*}

For ADE20K semantic segmentation, we follow the setting in~\cite{he2021masked} to fine-tune the pre-trained ViT models on this dataset. We always use a ViT with patch size $16$ and image size $512{\times}512$, which leads to a fixed sequence length of $L{=}1024$. We report the mean Intersection-over-Union (mIoU) averaged over 3 runs due to higher variance observed on ADED20K than COCO.

\paragraph{Baseline comparisons.} \cref{tab:baseline} shows the results along with the default parameters of our long-sequence MAE in comparison with previous works BEiT~\cite{Bao2021} and MAE~\cite{he2021masked}. While our pre-training cost is \app$4\x$ of MAE, BEiT (without dropping 75\% of the tokens) is also conceptually \app$4\x$ as high. It can be seen that our default model achieves notably higher performance on both tasks with a fixed fine-tuning budget. 

\paragraph{Ablation 1: $I$, $p$ and $L$.} We start out analyzing the key factors of our improved pre-training in \cref{tab:ablation_study}, top row. We adopt the methodology discussed in \cref{sec:approach:method} by fixing one and jointly varying two other input dimensions.
We first fix the image size $I$ in \cref{tab:ablate_image_size} or fix the patch size $p$ in \cref{tab:ablate_patch_size}. In both cases, performances still vary significantly -- suggesting sequence length is the key.
Indeed, when keeping $L=784$ as in \cref{tab:ablate_sequence_length}, all pre-trained models have similar performance and all notably outperform the MAE baseline in \cref{tab:baseline} -- this indicates that the sequence length $L$ itself -- as opposed to the image size $I$ or the patch size $p$ -- is the most important factor for the quality of pre-trained features.

\paragraph{Ablation 2: mask size $m$.} As described in \cref{sec:approach}, we adopt a joint masking strategy and decouple the mask size from the patch size. In \cref{tab:ablate_mask_size}, we find that joint masking with $m=2\times 2$ blocks on the feature grid works the best with a long sequence ($L=784$), which confirms our intuition in \cref{sec:approach:ours} that joint masking preserves task difficulty level when sequence length is increased. With $1\x 1$ masking, the results are almost at-par with MAE; with $4\x 4$ masking, the task may have become overly difficult and start to hurt.

\paragraph{Ablation 3: encoder/decoder sequence lengths.} Next, we study how the sequence length impacts individual components of MAE, in case further computational savings are desired. Specifically, we study $L_e$ and $L_d$ in \emph{isolation} (\cref{tab:ablate_enc_dec}).
Starting from our default setting in the middle row, we try decreasing encoder length $L_e$ in the first row by
employing a higher mask ratio of $\gamma = 0.9375$ ($49$ visible patches per image). We choose this ratio since it has the \emph{same} $L_e$ as the MAE baseline while matching our default decoder length. Interestingly, the result is even \emph{worse} than our baseline despite faster speed than our default setting. This means simply increasing mask ratio $\gamma$ does not directly result in better representations, and having decoupled mask size together with a longer sequence is a better solution.

We also experimented decreasing $L_d$ in \cref{tab:ablate_enc_dec}, last row, using a learned $2\times 2$ convolution that down-samples the full sequence from $28\times 28$ to $14\times 14$ after padding and before feeding it into the decoder. It also under-performs our default setting, but is notably better than decreasing $L_e$. These results show that both longer $L_e$ and longer $L_d$ help the feature quality, with $L_e$ being more important -- perhaps because the encoder is the one directly transferred to downstream tasks.

\paragraph{Ablation 4: longer pre-training.} Long-sequence MAE ($L{=}784$) increases the pre-training time by \app$4\times$ over the MAE baseline ($L{=}196$). This is reasonable since the dominating cost in ViT comes from the fully-connected layers, especially when the model size is large. Therefore, besides the BEiT baseline in \cref{tab:baseline} (also \app$4\x$), we further compare it to another setting in \cref{tab:ablate_4x_schedule} where a $4\times$ long learning schedule is used for original MAE (\ie a total number of 16,000 epochs on COCO). It can be seen that our long-sequence MAE significantly outperforms this MAE variant, which shows signals of overfitting on COCO and saturation on ADE20K. This suggests that it is better to spend additional computation by increasing sequence length $L$, than using it for a longer training schedule.

\paragraph{Is it merely because of larger transfer length?} One hypothesis on why long sequence can help is that the sequence length for downstream transfers ($4096$ for COCO; $1024$ for ADE20K) is significantly larger than MAE ($L{=}196$), and long-sequence MAE is just closing the gap. To see if it is the case, we perform an additional analysis on ADE20K with $384\x384$ input ($L{=}576$).\footnote{The COCO object detector heavily relies on pre-defined anchors~\cite{He2017} and other details that make it non-trivial to na\"{i}vely change the input size.} The baseline MAE gets $48.1$, at-par with $512\x512$ fine-tuning ($48.2$). Long-sequence MAE ($L{=}784$) is significantly better than this baseline, achieving $49.8$ despite a \emph{reduced} transferring sequence used after pre-training. This shows that long-sequence pre-training is generally beneficial beyond closing the length gap.

\begin{figure*}[t]
\vspace{-1.5em}
\centering
\begin{tabular}{cc}
\includegraphics[width=0.4\textwidth]{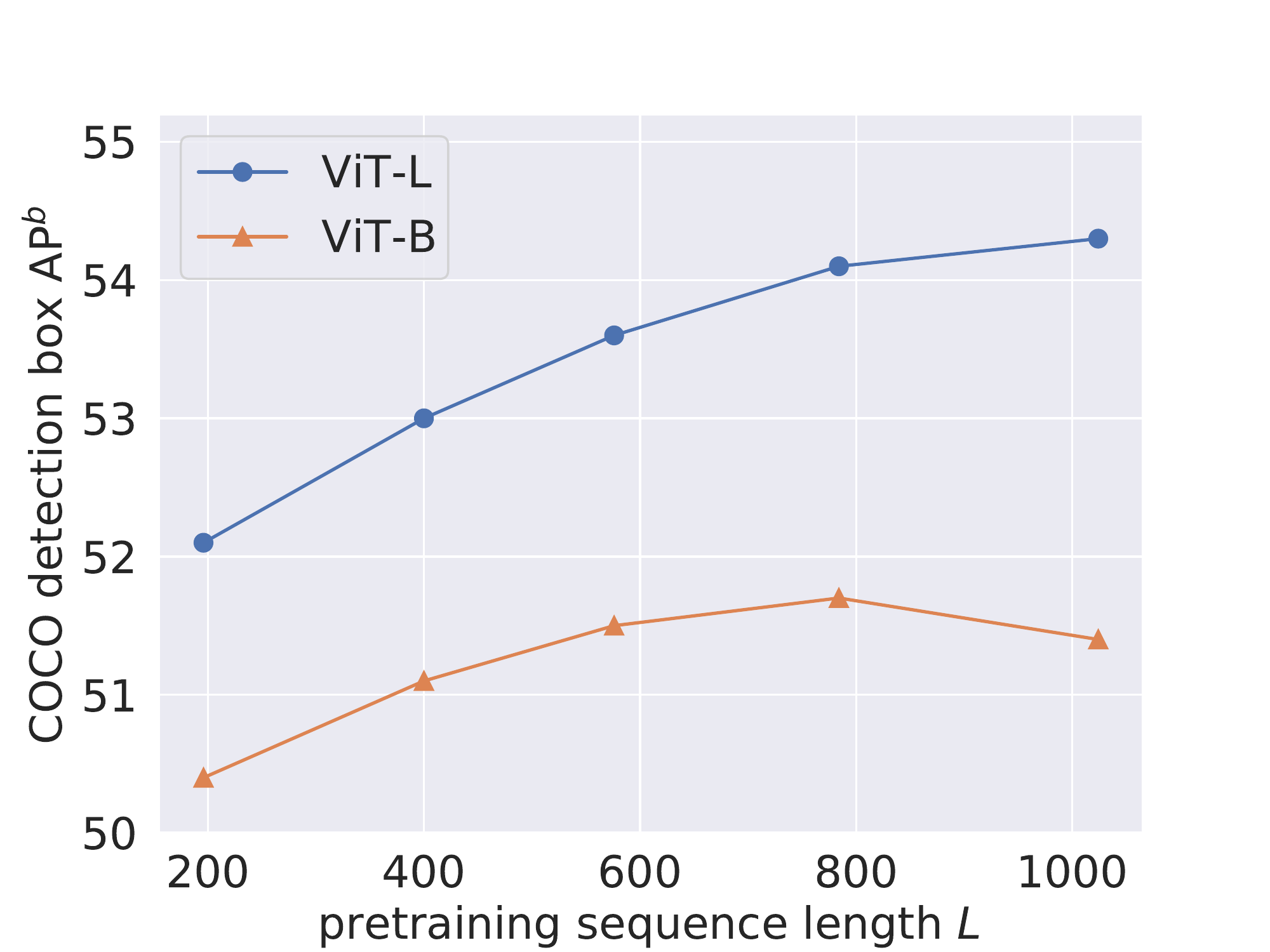} &
\includegraphics[width=0.4\textwidth]{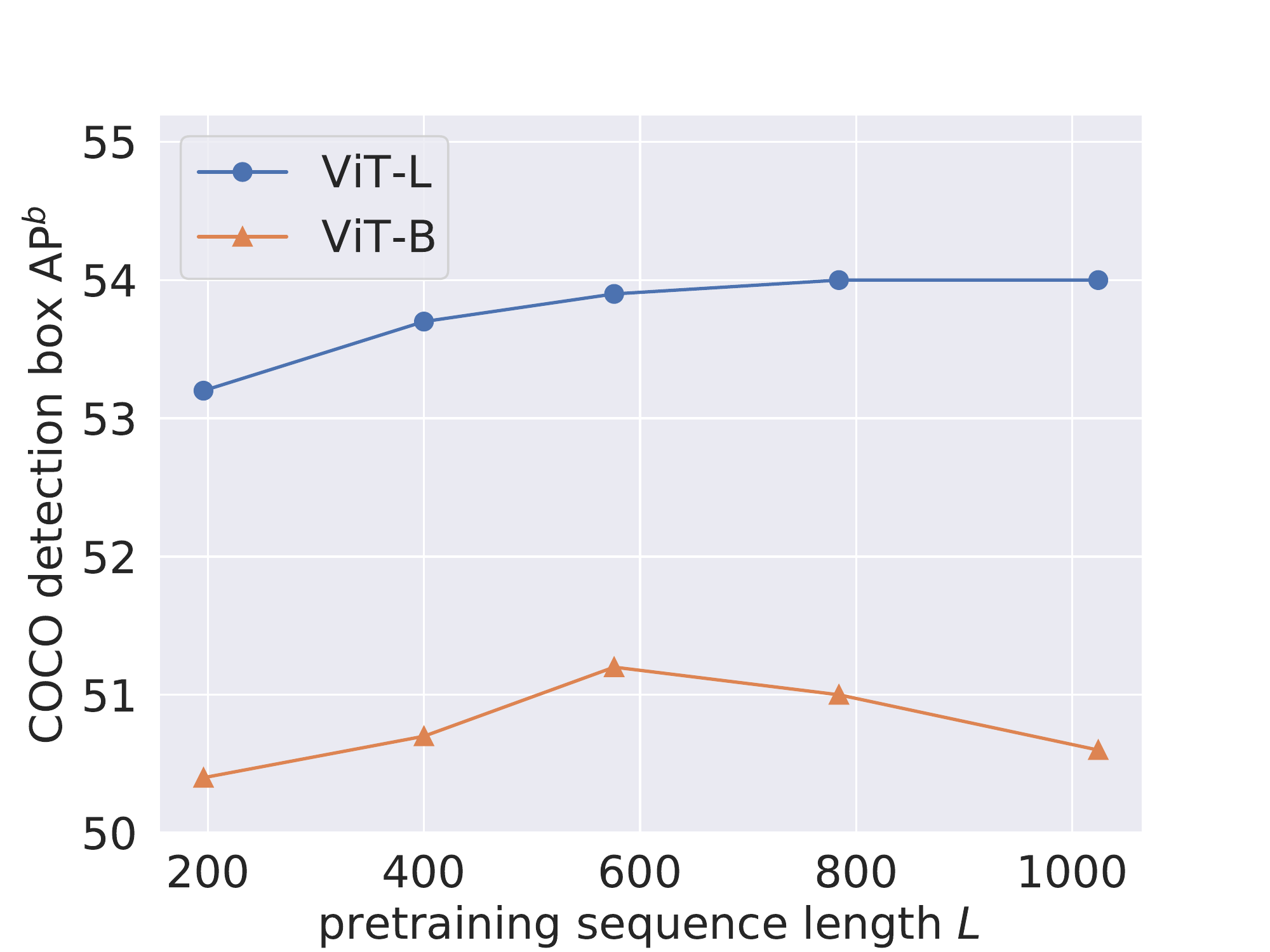} \\
a) pre-training on COCO &
b) pre-training on ImageNet-1K \\
\end{tabular}
\caption{\textbf{Scaling trends of sequence length $L$} for pre-training on COCO or ImageNet-1K; and evaluating on COCO object detection with fixed transferring length. We make three observations: 1) Increasing sequence length $L$ generally leads to a healthy trend of better AP$^{b}$ on COCO object detection. 2) The trend for ViT-L is better than ViT-B -- the former continues to scale even at $L{=}1024$, whereas the latter saturates and even starts to decline after a certain point. 3) COCO pre-training benefits more from longer sequences than ImageNet-1K.
\label{fig:seq_length_scaling}
}
\vspace{-.1in}
\end{figure*}

\subsection{Scaling trends of sequence lengths\label{sec:exp:seq_length}}

Given that sequence length $L{=}784$ improves downstream tasks as shown in \cref{sec:exp:ablation_study}, we next perform an in-depth study on the scaling trends of $L$. Specifically, we keep the patch size $p$ fixed as $16$, while varying the image size $I$ in $\left[224, 320, 384, 448, 512\right]$, corresponding to $L$ in $\left[196, 400, 576, 784, 1024\right]$. In addition to pre-training on COCO, we also include the ImageNet-1K dataset~\cite{Deng2009} as another pre-training source given its common use in previous work~\cite{he2021masked,xie2021simmim}. For ImageNet-1K, we follow the standard setting and pre-train on its \texttt{train} split with 1,281,167 images under an 800-epoch schedule~\cite{he2021masked}, which is roughly comparable to the total number of iterations in our COCO pre-training. To study how different ViTs behave with $L$, we also experiment with ViT-L~\cite{Dosovitskiy2021}, where we keep the same pre-training hyper-parameters as used in ViT-B (see \cref{tab:baseline}).

For each pre-trained model, we report results on the COCO object detection task. For ViT-L, we again fine-tune for 50-epochs on the COCO annotations and follow the corresponding ViT-L fine-tuning hyper-parameters in~\cite{li2021benchmarking}.

The results are shown in \cref{fig:seq_length_scaling}. Overall, increasing the pre-training sequence length $L$ from $196$ to $1024$ generally brings continued benefits to the transfer performance on the COCO detection box AP$^{b}$. This is especially clear for the larger-sized ViT-L, while for the smaller ViT-B, AP$^{b}$ saturates and even starts to decline after a certain sequence length for both COCO and ImageNet-1K pre-training. We suspect that ViT-L better scales to longer pre-training sequences than ViT-B because the former has a higher modeling capacity to handle a larger number of input image patches and learn rich features to represent the relationship among these patches (such as the part-whole relationship within an object or a scene, or the context co-occurrence relationship between visual components~\cite{Lin2014}). Given this observation, we hypothesize that it would be best to jointly scale the model parameters and the pre-training sequence length to learn stronger feature representations~\cite{tan2019efficientnet,radosavovic2020designing}.

In addition, comparing COCO pre-training in \cref{fig:seq_length_scaling}a) with ImageNet-1K pre-training in \cref{fig:seq_length_scaling}b), it can be seen that COCO pre-training benefits more from longer sequences: the former improves by more than $2$ AP points, whereas the latter by barely $1$ point. While not justified by controlled studies, we speculate this is because COCO images contain more objects and scene context on average than iconic ImageNet-1K images~\cite{Lin2014}, and are therefore more friendly to longer sequences that capture these objects and their interactions.

\subsection{More pre-training datasets\label{sec:exp:other_pre-training_datasets}}

\begin{table*}[t]
\centering
\tablestyle{7.5pt}{1.2}
\begin{tabular}{cl|ccc|ccc|ccc|ccc}
\multicolumn{2}{c|}{pre-training} & \multicolumn{3}{c|}{COCO~\cite{Lin2014}} & \multicolumn{3}{c|}{ImageNet-1K~\cite{Deng2009}} & \multicolumn{3}{c|}{Open Images~\cite{kuznetsova2020open}} & \multicolumn{3}{c}{Places~\cite{zhou2017places}} \\
\multicolumn{2}{c|}{dataset} & \multicolumn{3}{c|}{\emph{(241,690 images)}} & \multicolumn{3}{c|}{\emph{(1,281,167 images)}} & \multicolumn{3}{c|}{\emph{(1,743,042 images)}} & \multicolumn{3}{c}{\emph{(1,803,460 images)}} \\[.5mm]
\hline
encoder & method &
AP$^{b}$ & AP$^{m}$ & mIoU &  AP$^{b}$ & AP$^{m}$ & mIoU &  AP$^{b}$ & AP$^{m}$ & mIoU &  AP$^{b}$ & AP$^{m}$ & mIoU \Tstrut\\ \shline
\multirow{2}[0]{*}{ViT-B}
& baseline &
50.5 & 44.9 & 48.2 &  49.9 & 44.6 & 47.5 &  49.9 & 44.6 & 47.8 &  49.2 & 43.9 & 47.9 \\ & ours &
\textbf{51.7} & \textbf{45.9} & \textbf{50.8} &  \textbf{51.0} & \textbf{45.4} & \textbf{48.7} &  \textbf{51.0} & \textbf{45.4} & \textbf{49.8} &  \textbf{50.5} & \textbf{45.0} & \textbf{50.1} \\ \hline
\multirow{2}[0]{*}{ViT-L}
& baseline &
53.2 & 47.1 & 51.6 &  53.2 & 47.1 & 53.6 &  53.0 & 47.3 & 52.4 &  52.4 & 46.6 & 53.0 \\ & ours &
\textbf{54.1} & \textbf{48.0} & \textbf{54.6} &  \textbf{54.0} & \textbf{47.9} & \textbf{54.2} &  \textbf{54.1} & \textbf{47.9} & \textbf{54.7} &  \textbf{53.8} & \textbf{47.7} & \textbf{55.7} \\ \end{tabular}
\vspace{-.1em}
\caption{\textbf{More pre-training datasets} with the MAE baseline ($L{=}196$) and our long sequence MAE ($L{=}784$), evaluated with fixed fine-tuning length on COCO (AP$^{b}$, AP$^{m}$) and ADE20K (mIoU). A longer sequence during pre-training consistently benefits \textbf{all} the experimental settings across different pre-training datasets, downstream transfers, and model sizes. Different pre-training sources also behave differently in terms of data efficiencies. For example, we find COCO pre-training to be highly effective for our tasks of interest, despite having fewer images.
\label{tab:other_pre-training_datasets}}
\vspace{-.1in}
\end{table*}

Beyond COCO and ImageNet-1K, in this section we apply long-sequence MAE on more image sources for pre-training. Specifically, we choose two well-known datasets at a similar scale as ImageNet-1K, namely Open Images~\cite{kuznetsova2020open} and Places~\cite{zhou2017places}. For Open Images, we pre-train on its training split with 1,743,042 images. For Places, we pre-train on the Places365-Standard training split with 1,803,460 images from 365 scene categories. For both ViT-B and ViT-L, we pre-train the baseline MAE and our long-sequence MAE for 800 epochs on Open Images and 600 epochs on Places, respectively. All other hyper-parameters and details are kept the same as default (\cref{tab:baseline}).

Following \cref{sec:exp:ablation_study}, we evaluate all pre-trained models on COCO object detection, instance segmentation, and ADE20K semantic segmentation, where a fixed sequence length is used during fine-tuning for all evaluations. The results are summarized in \cref{tab:other_pre-training_datasets}. It can be seen that our long sequence pre-training ($L{=}784$) consistently outperforms the baseline setting ($L{=}196$) across \textbf{all} pre-training datasets and downstream tasks for both ViT-B and ViT-L. This confirms that long-sequence pre-training is a generic approach applicable and beneficial to various settings.

More interestingly, we also observe that different pre-training sources have different data efficiencies. Despite having only a fraction of images compared to other datasets (\eg, ${\app}1/5$ of ImageNet-1K), COCO pre-training is highly effective. For example, it achieves the \emph{highest} mIoU ($50.8$) for ViT-B among all pre-training sources when transferring to ADE20K semantic segmentation, as well as on-par or better COCO detection and ADE20K segmentation accuracies than ImageNet-1K pre-training.\footnote{We also tried varying the pre-training epochs on both datasets and found that the optimal COCO pre-trained models are consistently comparable to or better than the optimal ImageNet-1K ones for both ViT-B and ViT-L.} As discussed in \cref{sec:exp:seq_length}, we suspect this higher data efficiency is related to more scene-level images and a higher average number of objects in COCO, which benefits from a longer sequence. Furthermore, Places have a comparable number of images to Open Images, but the pre-trained ViT-L model on Places achieves the \emph{best} results on ADE20K with $55.7$ mIoU -- significantly outperforming the $53.3$ and $53.6$ mIoUs reported in BEiT~\cite{Bao2021} and MAE~\cite{he2021masked}. We attribute this to the rich variety of scenes in Places that helps learn better segmentation features and the close proximity in distribution to ADE20K~\cite{Zhou2019}.

\subsection{COCO and LVIS with SimpleFPN\label{sec:exp:simplefpn_coco_lvis}}

In this section, we further experiment with more downstream tasks and model architectures. We add LVIS~\cite{Gupta2019} as another benchmark, and evaluate the pre-trained encoders by fine-tuning them with the state-of-the-art SimpleFPN~\cite{li2022exploring} for plain ViT-based object detection and instance segmentation on both COCO and LVIS. Compared to the basic Mask R-CNN detector developed in~\cite{li2021benchmarking}, SimpleFPN \cite{li2022exploring} adopts a simpler and more efficient feature pyramid architecture~\cite{Lin2017}, adapts the pre-trained backbone for detection with residual convolutions~\cite{He2016}, and achieves a stronger performance.

Specifically, we pre-train the ViT models on both COCO and ImageNet-1K, using both the baseline MAE ($L=196$) and our long-sequence MAE ($L{=}784$). To match the setup in~\cite{li2022exploring}, we pre-train on ImageNet-1K for 1600 epochs. For COCO, we keep the 4,000-epoch schedule, as we find that more epochs do not give better performance. Then we fine-tune each pre-trained model separately on the COCO and LVIS bounding box and instance segmentation annotations with a fixed sequence length, and report their box AP$^{b}$ and mask AP$^{m}$ on both benchmarks.\footnote{For COCO, we follow the same hyper-parameters as in Table~5 of~\cite{li2022exploring}. On LVIS, we follow the hyper-parameters in Fig.~4 of~\cite{li2022exploring} (\eg $1024\times 1024$ image size), so that our LVIS results are comparable to Fig.~4 of~\cite{li2022exploring}.}

The results are shown in \cref{tab:simplefpn_coco_lvis}, where models pre-trained with an increased sequence length again consistently outperform the baseline MAEs under both COCO and LVIS APs across different encoder architectures, despite all fine-tuning hyper-parameters are optimized for the baseline. In fact, the gains on LVIS are \emph{more salient} than those on COCO, for example $+2.9$ in box AP$^{b}$ from $43.3$ to $46.2$ with ViT-L pre-trained on COCO. When checking the detailed breakdown for classes with different frequencies, we find that long-sequence MAE helps more for long-tail classes. This further enhances our conclusion that long-sequence pre-training helps, this time generalizing to a new downstream task (LVIS) and the latest detector architecture (SimpleFPN).

\begin{table*}[t]
\centering
\tablestyle{7.5pt}{1.2}
\begin{tabular}{cl|cccc|cccc}
\multicolumn{2}{c|}{pre-training dataset} & \multicolumn{4}{c|}{COCO~\cite{Lin2014}} & \multicolumn{4}{c}{ImageNet-1K~\cite{Deng2009}} \\[0.75mm]
\hline
encoder & method & AP$^{b}_{coco}$ & AP$^{m}_{coco}$ & AP$^{b}_{lvis}$ & AP$^{m}_{lvis}$ & AP$^{b}_{coco}$ & AP$^{m}_{coco}$ & AP$^{b}_{lvis}$ & AP$^{m}_{lvis}$ \Tstrut\\[1mm]
\shline
\multirow{3}[0]{*}{ViT-B}
& SimpleFPN~\cite{li2022exploring} & -- & -- & -- & -- & 51.6 & 45.9 & 40.2 & 38.2 \Tstrut\\
& baseline &
51.9 & 46.1 & 39.6 & 37.4 &  51.8 & 46.0 & 40.3 & 38.1 \\ & ours &
\textbf{52.5} & \textbf{46.7} & \textbf{41.6} & \textbf{39.1} &  \textbf{52.1} & \textbf{46.2} & \textbf{40.8} & \textbf{38.5} \\
\hline
\multirow{3}[0]{*}{ViT-L}
& SimpleFPN~\cite{li2022exploring} & -- & -- & -- & -- & 55.6 & 49.2 & 46.0 & 43.4 \Tstrut\\
& baseline &
54.4 & 48.3 & 43.3 & 41.0 &  55.3 & 49.0 & 45.2 & 42.7 \\ & ours &
\textbf{56.0} & \textbf{49.5} & \textbf{46.2} & \textbf{43.5} &  \textbf{56.1} & \textbf{49.7} & \textbf{46.6} & \textbf{44.0} \\ \end{tabular}
\vspace{-.1em}
\caption{\textbf{COCO and LVIS with SimpleFPN} for object detection and instance segmentation.
For each model size (ViT-B and ViT-L), the first row contains the box AP$^{b}$ and mask AP$^{m}$ as reported in~\cite{li2022exploring}, for which the same pre-training recipe is used in our baseline. The second and third rows are results from our experiments. Our baseline can reproduce the results of~\cite{li2022exploring} while our long-sequence MAE consistently outperforms the baseline, especially on LVIS.}
\label{tab:simplefpn_coco_lvis}
\vspace{-.1in}
\end{table*}

\begin{table}[t]
\centering
\tablestyle{5pt}{1.2}
\begin{tabular}{x{30}y{30}|x{50}x{60}}
\multicolumn{2}{c|}{pre-training dataset} & COCO~\cite{Lin2014} & ImageNet-1K~\cite{Deng2009} \\[0.75mm]
\shline
\multirow{2}[0]{*}{ViT-B}
& baseline & 83.2 & \textbf{83.6} \\
& ours & \textbf{83.8} & 83.5 \\
\hline
\multirow{2}[0]{*}{ViT-L}
& baseline & 84.9 & \textbf{85.9} \\
& ours & \textbf{85.5} & 85.7 \\
\end{tabular}
\vspace{-.1em}
\caption{We use \textbf{ImageNet-1K classification} as an additional downstream evaluation to complete our assessment of long-sequence pre-training for MAE. Unlike detection or segmentation tasks, the signal for classification is mixed, with noticeable gains from COCO pre-training, and no further benefit from ImageNet-1K. The same trend is observed for both ViT-B and ViT-L.
\label{tab:imagenet_classification}}
\vspace{-.1in}
\end{table}

\subsection{ImageNet-1K classification\label{sec:exp:in1k_ft}}

So far we have mainly focused on the localization tasks that operate on scene-level images: object detection, instance segmentation, and semantic segmentation. To give a more complete picture of our long-sequence pre-training, we next extend the downstream evaluation to standard image classification on the ImageNet-1K dataset~\cite{Deng2009}.

We conduct two sets of experiments. First, we use the standard classification setting and fix the input sequence length during fine-tuning ($I{=}224$, $p{=}16$, and $L{=}196$), given pre-trained models either from COCO (4,000-epoch) or ImageNet-1K (1600-epoch). Following the practice from previous sections, the pre-training is done either with or without long-sequence inputs, and for both ViT-B and ViT-L. Fine-tuning hyper-parameters and details strictly follow~\cite{he2021masked}. The results (in top-1 accuracy) are summarized in \cref{tab:imagenet_classification}. Interestingly, the signal is \emph{mixed} and is dependent on the pre-training dataset. With COCO, long-sequence pre-training offers substantial gains despite a reduced sequence length during supervised classification. In fact, for ViT-B, long-sequence MAE from COCO gives better ImageNet-1K accuracy ($83.8\%$) than any of our ImageNet-1K pre-trained MAEs, while using merely ${\app}1/5$ the number of images. On the other hand, ImageNet-1K pre-training with longer sequences does not bring immediate benefits for ImageNet classification if the fine-tuning length is fixed.

However, the above results do not mean long-sequence is not useful when dealing with ImageNet-1K images alone. On the contrary, we find sequence length is still playing a vital role in further driving the top-1 accuracy. In MAE, a state-of-the-art result ($87.8\%$) is achieved with input size $I=448$~\cite{he2021masked}, which effectively adopts a total sequence length of $L=1024$. To show that $L$ is more important than $I$, we pre-train a ViT-H model on ImageNet-1K for 800 epochs, with image size $I=224$ and patch size $p=7$, which also arrives at a total sequence length of $L{=}1024$. 
After pre-training, we fine-tune the model on ImageNet-1K with classification labels, following the ViT-H hyper-parameters and other details in~\cite{he2021masked}. For fine-tuning, we also stick to the image size $I=224$ and patch size $p=7$, and hence the same sequence length $L=1024$.

We evaluate the resulting model with standard $224\x 224$ center-crop testing, and it gives a top-1 accuracy of $87.7\%$. This is comparable to MAE's best result with the same model size, and is achieved with the same sequence length $L$, but \emph{without} using a larger image size (or additional peripheral pixels~\cite{Touvron2021a}). This again shows that sequence length is the key to the performance boost, for both self-supervised pre-training and supervised classification as mentioned in~\cite{beyer2022better}.

\section{Conclusion\label{sec:conclusion}}

In this work, we have explored long-sequence MAE pre-training, which has shown consistent improvements across various pre-training datasets and downstream benchmarks.
The encouraging results suggest that sequence length is a viable axis for scaling, and has a potential \emph{compound effect}~\cite{tan2019efficientnet} with the type of data used (\eg scene-level~\cite{Lin2014,zhou2017places} \vs object-level~\cite{Deng2009}) and other important axes like model size.
In contrast to model size, longer sequences during pre-training do \emph{not} necessarily imply longer sequences during transferring, and indeed a \emph{fixed} input size for evaluations is the analysis protocol we rigorously followed. 

We have also established multiple MAE baselines on other image sources beyond ImageNet-1K, and showed that they can provide equivalent or better data efficiency and feature quality for relevant tasks. We hope this expansion can provide a more complete assessment of MAE.

One potential limitation of our work is that an increased sequence length inevitably increases the computational cost and thus the carbon footprint during pre-training. However, we believe this cost can be amortized by the various number of application possibilities from a single pre-training run; and justified by the performance gains without incurring extra costs during such transfers. While our current findings are mostly empirical, we hope they can aid future theoretical explanations and inspire more studies on this frontier.

\paragraph{Acknowledgments.} We are grateful to Ross Girshick, Kaiming He, and Alex Berg for helpful discussions on various ideas and experiments, and Yanghao Li and Hanzi Mao for their code base and examples of COCO and LVIS instance segmentation experiments before the public release, as well as Alexander Kirillov for the support on exploring other data sources for pre-training. We thank the Google TPU team for their Cloud TPU support.

\appendix

\begin{table*}[t]
\centering
\tablestyle{15pt}{1.2}
\begin{tabular}{cl|ccc|ccc}
\multicolumn{2}{c|}{pre-training splits} & \multicolumn{3}{c|}{\texttt{train2017}} & \multicolumn{3}{c}{\texttt{train+unlabeled2017}}\\
\multicolumn{2}{c|}{(COCO)} & \multicolumn{3}{c|}{\emph{(118,287 images)}} & \multicolumn{3}{c}{\emph{(241,690 images)}} \\[.5mm]
\hline
encoder & method &
AP$^{b}$ & AP$^{m}$ & mIoU &  AP$^{b}$ & AP$^{m}$ & mIoU \Tstrut\\ \shline
\multirow{2}[0]{*}{ViT-B}
& baseline &
49.7 & 44.1 & 46.9 &  50.5 & 44.9 & 48.2 \\ & ours &
\textbf{51.0} & \textbf{45.3} & \textbf{49.5} &  \textbf{51.7} & \textbf{45.9} & \textbf{50.8} \\ \hline
\multirow{2}[0]{*}{ViT-L}
& baseline &
50.7 & 44.9 & 47.9 &  53.2 & 47.1 & 51.6 \\ & ours &
\textbf{52.3} & \textbf{46.4} & \textbf{51.8} &  \textbf{54.1} & \textbf{48.0} & \textbf{54.6} \\ \end{tabular}
\vspace{-.1em}
\caption{Images from the COCO \texttt{unlabeled2017} split roughly double the pre-training data size and greatly help downstream transfers to both COCO detection (AP$^{b}$, AP$^{m}$) and ADE20K segmentation (mIoU), evaluated under the same setting as in \cref{tab:other_pre-training_datasets}. We pre-train for 8,000 epochs on \texttt{train2017} and 4,000 epochs on (\texttt{train+unlabeled2017}), which share roughly the same number of iterations and are near optimal for both pre-training data settings.
\label{tab:supp_coco_splits}
\vspace{-.1in}}
\end{table*}

\section{Implementation details\label{sec:appendix}}

\paragraph{Pre-training details.} We build our long-sequence pre-training upon the public MAE open-source repository implemented in PyTorch\footnote{\url{https://github.com/facebookresearch/mae}} and adapt it to run on Cloud TPUs using PyTorch/XLA\footnote{\url{https://pytorch.org/xla}} for our pre-training experiments. Our long-sequence pre-training mostly involves changing the image size $I$ and the patch size $p$ in the data loader and ViT model definition in this code base, as well as changing the sampling procedure for masked patch indices following our joint $m=2\times 2$ decoupled masking. Specifically, we move the mask index generation from the MAE network to the data loader to easily apply different masking strategies.

We follow the implementation details in the MAE open-source repository mentioned above in our pre-training experiments, except that we use a slightly smaller decoder (hidden size $384$ instead of $512$, and number of heads $12$ instead of $16$) in our ViT-B experiments throughout the paper for both the baseline MAE setting ($L=196$) and our long-sequence setting ($L=784$). This choice has a historical reason: In the early stage of the project, ViT-S (hidden size $384$) is among the encoders we used for efficient research explorations. For such a small model, maintaining a $512$ dimensional decoder \emph{offsets} the asymmetric and efficient design of MAE as it leads to a \emph{heavier} decoder that processes all the tokens. Therefore, our MAE design always keeps the decoder dimension \emph{half} of the encoder dimension, and keeps the same number of heads to the encoder, for better adaptation to various-sized encoders. For ViT-B, it leads to a hidden size of $384$ and $12$ heads; for ViT-L, it is $512$ (same as the original MAE).

When reducing the decoder sequence length $L_d$ in \cref{tab:ablate_enc_dec} row 3, we insert a learned $2\times 2$ convolutional layer with stride 2 before the first decoder ViT block to down-sample $L_d$ from $28\times 28$ to $14 \times 14$, which we find is slightly better than average pooling. In this case, we also reshape the target image (in the MAE reconstruction $\ell_2$ loss) to $14 \times 14$ patches with doubled patch size $2p$ to matched the down-sampled $L_d$ and train the decoder to predict $2p \times 2p \times 3$ pixel targets.

\paragraph{Dataset details.} In our pre-training experiments on COCO, we find that pre-training on images from the union of \texttt{train2017} and \texttt{unlabeled2017} splits greatly improves the results over pre-training on \texttt{train2017} alone, as detailed in \cref{tab:supp_coco_splits}. Notably, the former setting doubles the pre-training data size from 118,287 to 241,690. In addition to better transferring performance to ADE20K, the downstream COCO object detection and instance segmentation task itself (exclusively trained on the annotations from COCO \texttt{train2017}) also largely benefits from pre-training with the \texttt{unlabeled2017} split. To our knowledge, this is one of the \emph{first} use cases of the COCO unlabeled images that shows promising signals for object detection and segmentation.
This confirms that the MAE pre-training helps downstream tasks through learning scalable (\wrt dataset size), unsupervised visual representations.

In our Open Images pre-training, since the dataset contains many high-resolution images, we first resize all the images to a long side of 640 pixels (following the image size of COCO) as a pre-processing step to reduce data loading overhead during pre-training. We use bicubic interpolation from the PIL library and save the resized images with a JPEG quality of 95, which gives better pre-training results than the default JPEG quality of 75 in the PIL library. When pre-training on all other datasets, we directly use the image files provided by these datasets.

\newpage

{\small
\bibliographystyle{ieee_fullname}
\bibliography{long_seq_mae}
}

\end{document}